\documentclass[10pt,twocolumn,letterpaper]{article}

\usepackage{cvpr}              

\usepackage{multirow}
\usepackage{placeins}
\definecolor{tablebestbg}{HTML}{B6D3EF}
\definecolor{tablesecondbg}{HTML}{E6F0FA}
\DeclareRobustCommand{\tablebest}[1]{\begingroup\setlength{\fboxsep}{1pt}\hspace{-1pt}\colorbox{tablebestbg}{\textbf{#1}}\hspace{-1pt}\endgroup}
\DeclareRobustCommand{\tablesecond}[1]{\begingroup\setlength{\fboxsep}{1pt}\hspace{-1pt}\colorbox{tablesecondbg}{\underline{#1}}\hspace{-1pt}\endgroup}
\newcommand{\R}{\mathbb{R}}
\newcommand{\p}{\mathbf p}

\definecolor{cvprblue}{rgb}{0.21,0.49,0.74}
\usepackage[pagebackref,breaklinks,colorlinks,allcolors=cvprblue]{hyperref}

\def\confName{CVPR}
\def\confYear{2027}

\title{SpectralGS: Spectral Gaussian Splatting for Unsupervised Spectral Super-Resolution}

\author{%
	Qiaoying Qu\textsuperscript{1}\quad Bin Pan\textsuperscript{1,*}\quad Chen Jia\textsuperscript{2}\quad Haiyang Xu\textsuperscript{1}\quad Xia Xu\textsuperscript{3}\quad Zhenwei Shi\textsuperscript{4}\\[6pt]
	{\small\textsuperscript{1}School of Statistics and Data Science, LEBPS, KLMDASR, LPMC, AAIS, NITFID, Nankai University}\\[2pt]
	{\small\textsuperscript{2}The Key Laboratory of Computer Vision and System, Ministry of Education, Tianjin University of Technology}\\[2pt]
	{\small\textsuperscript{3}School of Computer Science and Technology, Tiangong University}\\[2pt]
	{\small\textsuperscript{4}Department of Aerospace Intelligent Science and Technology, School of Astronautics, Beihang University}
}

\begin{document}
\maketitle
\begin{abstract}

Unsupervised spectral super-resolution aims to recover spectral information from RGB images without paired hyperspectral data. However, the limited spectral information in RGB observations makes learning an effective spatial–spectral representation challenging, requiring the incorporation of structured spectral modeling with sensor imaging. In this paper, we propose SpectralGS, a spectral Gaussian splatting framework comprising two components: Shared Spectral Gaussian Representation and Spectral Response Projection Rendering. The representation module constructs a continuous spatial-spectral representation via shared spectral and local spatial Gaussian primitives, exploiting the low-rank spectral structure of hyperspectral images. The rendering module utilizes the spectral response function (SRF) to apply integral projection to the continuous representation, blending it to generate an RGB image. Consistency between the rendered and observed RGB images provides training supervision. Our main contribution is a Gaussian splatting framework that integrates structured continuous spatial–spectral representation with imaging response-based rendering, enabling unsupervised spectral reconstruction. Extensive experiments demonstrate that our method outperforms current state-of-the-art methods.

\end{abstract}

\section{Introduction}

\begin{figure}[t]
\centering
\includegraphics[width=\columnwidth]{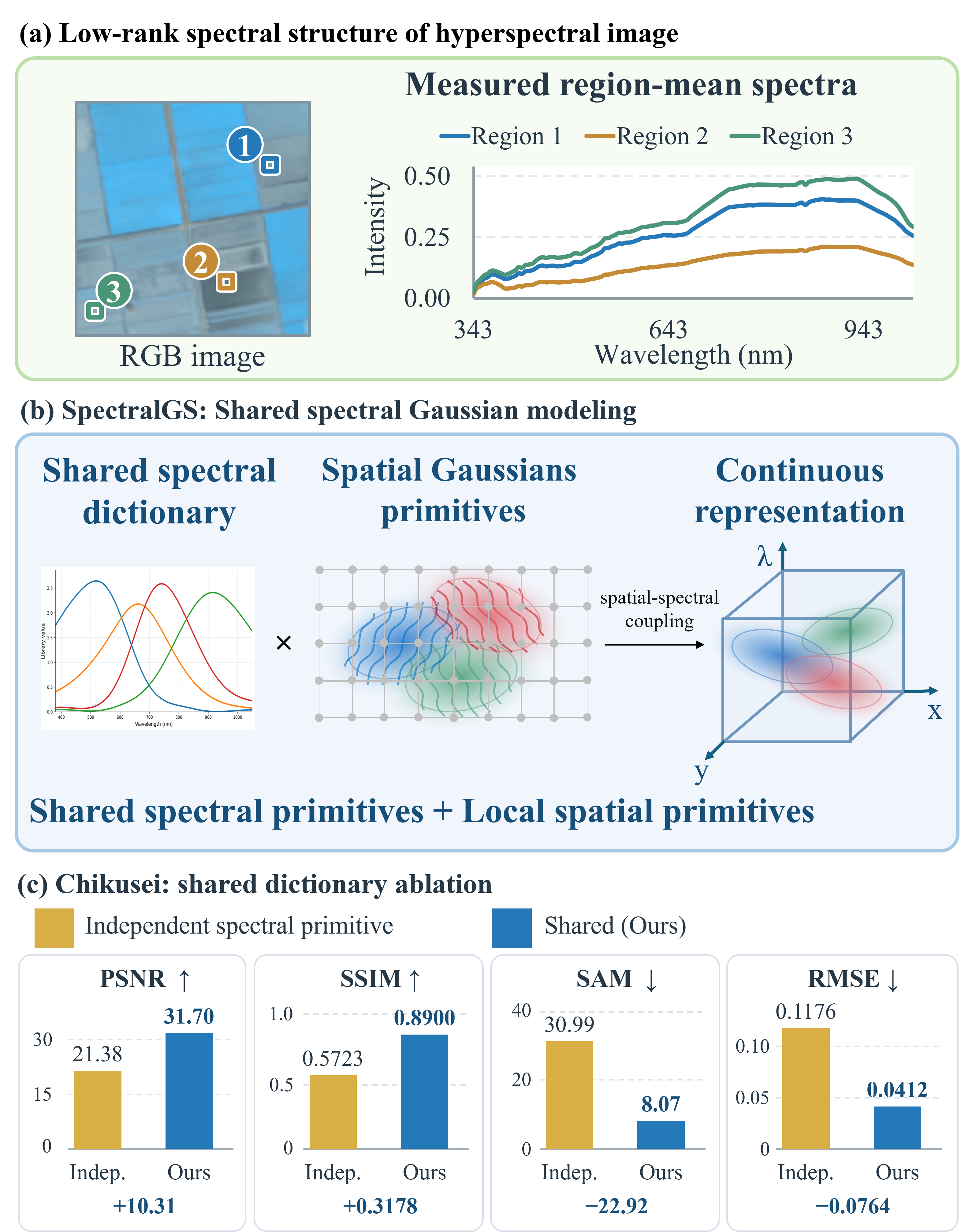}
\caption{Motivation and effect of shared spectral modeling. (a) Region-mean spectra from a representative HSI exhibit correlated shapes, motivating a shared spectral dictionary. (b) SpectralGS couples shared spectral primitives with local spatial primitives to form a continuous spatial-spectral representation. (c) On Chikusei, the shared dictionary consistently outperforms independent spectral primitives.}
\label{motivation}
\end{figure}

Hyperspectral images (HSIs) capture spatial information together with detailed spectral signatures, supporting applications such as classification~\cite{11367376,11417245,10696913}, target detection~\cite{11270006,11345538,11112647}. However, acquiring HSIs with high spatial resolution remains challenging because of hardware costs and imaging constraints. Spectral super-resolution (SSR) offers a computational alternative by reconstructing spectral information from multispectral or RGB images.

Supervised SSR methods aim to learn the mapping between RGB/multispectral observation and real hyperspectral image. Methods based on spectral priors constrain plausible reconstruction results by learning dictionaries, sparse representations, and Gaussian process models~\cite{10.1007/978-3-319-46478-7_2,8410422,8481553}. Convolutional methods capture local spatial-spectral dependencies through residual connections, multiscale aggregation, and channel recalibration~\cite{8575292,9150643,9857076}. Attention mechanisms, state-space, and frequency-domain methods are developed to capture longer-range dependencies or frequency information~\cite{Cai2022MSTMS,10887701,10494781}. Optimization-inspired algorithms combine learned prior information with spectral degradation~\cite{9357488,WANG2026114641,Young_2026_CVPR}. Structured representations further leverage low-rank features~\cite{10465659} or wavelength-conditional reconstruction~\cite{xu2024continuous,zhang2026radiativestructuredneuraloperatorcontinuous}. Despite these advances, supervised learning methods still require reference HSI data, which is often difficult to obtain for new sensors or scene domains.

Without paired hyperspectral images, unsupervised SSR methods reconstruct spectral information using various constraints. Basis-based and regularization-based approaches constrain reconstruction using spectral representations and priors~\cite{9150656,Zhang:21,jiang2026dmmaf}. Masked modeling captures inter-band correlations and spectral continuity~\cite{10093998}. Uncertainty-guided methods adaptively extract structural and semantic priors from RGB observations~\cite{10843147}. Adversarial approaches learn spectral distributions from unpaired data, optionally incorporating semantic constraints~\cite{9710095,9423159}. Unmixing-based methods extract spectral and spatial priors through endmember-abundance decomposition~\cite{10198468}. Sampling-guided approaches combine sampled spectra with RGB features to guide reconstruction~\cite{11098941}.

The limited spectral information available under RGB-only supervision makes it difficult to learn an explicit continuous spatial–spectral representation that captures spatial detail and spectral structure while remaining consistent with sensor imaging process. 3D Gaussian Splatting (3DGS)~\cite{10.1145/3592433} represents scenes using explicit Gaussian primitives and optimizes them through differentiable image rendering. Gaussian image representation further demonstrates how local Gaussian contributions can be accumulated to reconstruct images~\cite{10.1007/978-3-031-72673-6_18}. These properties motivate us to adapt Gaussian splatting to the spatial–spectral domain, where Gaussian primitives describe spatial and spectral information, and response-based rendering connects the representation to RGB observations. 

Directly extending 3DGS to SSR requires independently predicting numerous local spatial-spectral parameters, making optimization difficult. Inspired by the low-rank structure of hyperspectral images (Fig.~\ref{motivation} (a)), we propose SpectralGS, which incorporates spectral structure into continuous spatial–spectral modeling through a shared spectral dictionary. The framework consists of two components: Shared Spectral Gaussian Representation and Spectral Response Projection Rendering. In the representation module, shared spectral primitives form a spectral dictionary that captures common spectral structure across the image. Each local spatial Gaussian primitive combines the spectral primitives through learnable mixture weights to form a continuous spatial-spectral representation (Fig.~\ref{motivation}(b)). The rendering module conducts SRF integral projection on the shared spectral primitives to derive their RGB responses, which are then multiplied by the corresponding spatial components to determine the projected primitives. An RGB image is subsequently generated through sampling and additive blending. The consistency between the rendered RGB image and the observed RGB image provides the training constraint. Fig.~\ref{motivation} (c) shows that the shared dictionary improves all four metrics over independent spectra.

The major contributions of our algorithm can be summarized as follows:

\begin{itemize}
\item We propose a spectral Gaussian splatting framework for unsupervised SSR, which explicitly integrates a structured continuous spatial-spectral representation with the sensor imaging process, enabling spectral reconstruction under RGB-only supervision.

\item We develop Shared Spectral Gaussian Representation, which constructs a continuous representation incorporating shared spectral Gaussian primitives with local spatial Gaussian primitives, exploiting the low-rank spectral structure to constrain spectral modeling.

\item We introduce Spectral Response Projection Rendering, which produces RGB images from a continuous representation via SRF integral projection and additive blending, enabling unsupervised training through RGB rendering consistency.

\end{itemize}

\begin{figure*}[t]
	\centering
	\includegraphics[width=\textwidth]{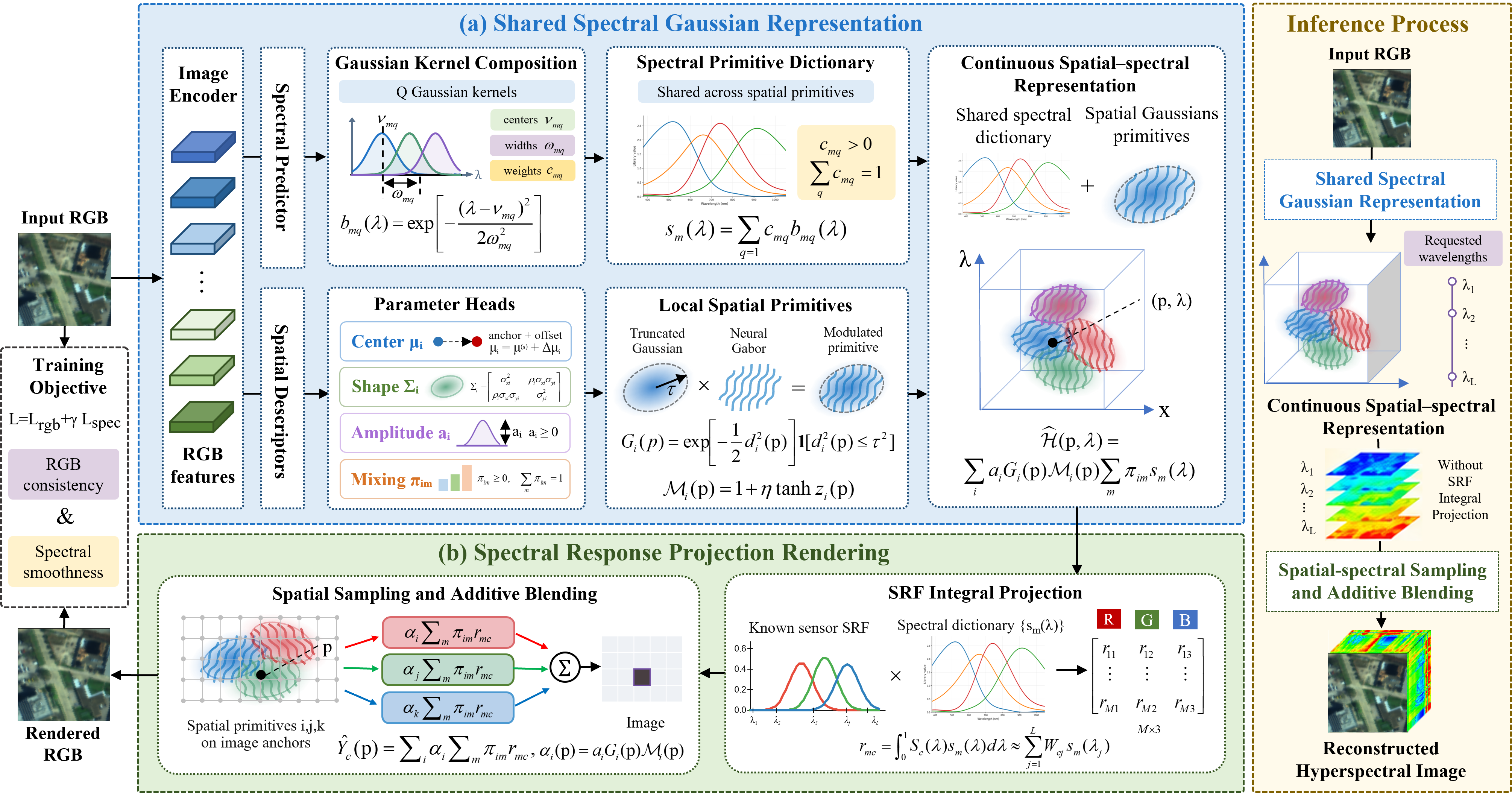}
	\caption{\textbf{Overview of SpectralGS}, including Shared Spectral Gaussian Representation and Spectral Response Projection Rendering for RGB-only training. The inference process for hyperspectral reconstruction is shown on the right.}
	\label{overall}
\end{figure*}

\section{Related work}
\subsection{3D Gaussian Splatting}
3D Gaussian Splatting (3DGS) represents scenes using explicit Gaussian primitives and synthesizes novel views through camera projection and alpha blending~\cite{10.1145/3592433}. Building on this framework, Mip-Splatting introduces frequency-aware filtering to improve rendering across different scales~\cite{10656976}, while Scaffold-GS organizes local Gaussians around anchors and predicts their attributes according to the viewing conditions~\cite{Lu_2024_CVPR}. Neural Gabor Splatting further extends the expressive capacity of individual primitives through local modulation, allowing finer surface details to be represented~\cite{Watanabe_2026_CVPR}. These developments improve Gaussian representations through scale-aware rendering, structured primitive organization, and richer local appearance modeling.

\subsection{Gaussian Splatting for Image Representation}
Gaussian splatting has also been adapted to 2D image signals. For image representation and compression, GaussianImage and GaussianImage++ represent images with spatial Gaussian primitives, combining efficient image synthesis with quantization and adaptive primitive allocation~\cite{10.1007/978-3-031-72673-6_18,li2026gaussianimagepp}. For arbitrary-scale spatial super-resolution, GaussianSR and GSASR learn continuous Gaussian representations from low-resolution images and use spatial rendering to reconstruct images at requested scales~\cite{hu2024gaussiansrhighfidelity2d,11446149}. These studies establish Gaussian splatting as an effective framework for image representation and reconstruction.

\section{Method}
This section presents the representation and rendering modules of SpectralGS, followed by its training objective, inference procedure, and optional spectral refinement.

\subsection{Problem formulation}
\label{problem}

Given an RGB image $Y\in\R^{H\times W\times3}$, spectral super-resolution aims to reconstruct an HSI $\widehat H\in\R^{H\times W\times L}$, where $H$ and $W$ are the spatial dimensions and $L$ is the number of spectral bands. We consider unsupervised training, where paired HSI targets are unavailable. Throughout this paper, $\lambda\in[0,1]$ denotes wavelength normalized over the modeled spectral range.

We describe the hyperspectral signal as a continuous function $\mathcal H(\p,\lambda)$, which gives the spectral intensity at pixel position $\p=(x,y)$ and wavelength $\lambda$. The RGB observation is obtained by integrating this function against the SRF:
\begin{equation}
	Y_c(\p)=\int_0^1
	S_c(\lambda)\mathcal H(\p,\lambda)\,d\lambda+e_c(\p),
	\label{observation}
\end{equation}
where $c\in\{R,G,B\}$ indexes RGB channels, $e_c(\p)$ denotes observation errors, and the known SRF satisfies $S_c\ge0$ and $\int_0^1S_c(\lambda)\,d\lambda=1$. 

\subsection{Overall architecture}
\label{overall_architecture}
As shown in Fig.~\ref{overall}, given the input RGB image $Y$, a predictor and a descriptor estimate the spectral dictionary parameters $\Theta_{\mathrm{spec}}$ and spatial primitive parameters $\Theta_{\mathrm{spatial}}$ respectively. Together, $\Theta=\{\Theta_{\mathrm{spec}},\Theta_{\mathrm{spatial}}\}$ defines the continuous representation:
\begin{equation}
	\widehat{\mathcal H}_{\Theta}(\p,\lambda)
	=\sum_{i=1}^{N}a_iG_i(\p)\mathcal M_i(\p)
	\sum_{m=1}^{M}\pi_{im}s_m(\lambda),
	\label{representation_flow}
\end{equation}
where $N$ and $M$ denote the numbers of spatial primitives and shared spectral primitives, respectively. For spatial primitive $i$, $a_i$ is its amplitude, $G_i(\p)$ its truncated Gaussian weight, and $\mathcal M_i(\p)$ its neural Gabor modulation. The function $s_m(\lambda)$ describes spectral primitive $m$, and $\pi_{im}$ is its mixture weight for spatial primitive $i$. These components are detailed in Sec.~\ref{representation}.

Spectral Response Projection Rendering maps this representation to RGB through SRF integral projection:
\begin{equation}
 \widehat Y_c(\p)=\int_0^1S_c(\lambda)
 \widehat{\mathcal H}_{\Theta}(\p,\lambda)\,d\lambda.
 \label{rendering_projection}
\end{equation}
The renderer evaluates the spectral integrals before spatial sampling and additive blending (Sec.~\ref{rendering}). RGB rendering consistency and spectral regularization guide training (Sec.~\ref{unsupervised_loss}). At inference, the representation is sampled directly at the requested wavelengths:
\begin{equation}
 \widehat H_l(\p)=\widehat{\mathcal H}_{\Theta}(\p,\lambda_l),
 \qquad l=1,\ldots,L.
 \label{reconstruction_flow}
\end{equation}
This reconstructs the HSI without SRF integral projection (Sec.~\ref{inference}). The optional post-hoc refinement is described separately in Sec.~\ref{sgre-method}.

\subsection{Shared Spectral Gaussian Representation}
\label{representation}
This section details the spectral dictionary, local spatial primitives and their joint continuous representation. We omit the parameter subscript $\Theta$ when it is fixed.

\subsubsection{Spectral dictionary via Gaussian Kernel Composition}

Hyperspectral images often exhibit a low-rank spectral structure due to strong correlations among spectral bands. Motivated by this property, we introduce a shared spectral dictionary to reduce redundant modeling.

For each input image, the predictor constructs a dictionary of $M$ shared spectral Gaussian primitives. Each primitive is synthesized through Gaussian Kernel Composition of $Q$ 1D Gaussian kernels:
\begin{equation}
 b_{mq}(\lambda)=\exp\left[-\frac{(\lambda-\nu_{mq})^2}{2w_{mq}^2}\right],
 \label{spectral_kernel}
\end{equation}
\begin{equation}
 \begin{gathered}
 \widetilde s_m(\lambda)=\sum_{q=1}^Q c_{mq}b_{mq}(\lambda),\\
 c_{mq}>0,\qquad \sum_q c_{mq}=1,
 \end{gathered}
 \label{factor}
\end{equation}
where $\nu_{mq}\in(0,1)$, $w_{mq}>0$, and $c_{mq}$ are the center, width, and composition coefficient of Gaussian kernel $q$ in spectral primitive $m$, respectively. We normalize each spectral primitive to unit area:
\begin{equation}
 s_m(\lambda)=\frac{\widetilde s_m(\lambda)}{Z_m},\qquad Z_m=\int_0^1\widetilde s_m(\lambda')\,d\lambda'.
 \label{library}
\end{equation}
This fixes the spectral scale, with local spatial amplitudes controlling the intensity.

\subsubsection{Local spatial primitives and neural Gabor modulation}
We place an anchor at each pixel location as a fixed reference position and associate it with $K$ spatial Gaussian primitives, giving $N=HWK$ primitives. Learned offsets determine their centers relative to the anchors. 

Each local primitive has a spatial center $\boldsymbol\mu_i^{xy}$, an amplitude $a_i\ge0$, and a covariance
\begin{equation}
\begin{gathered}
 \Sigma_i^{xy}=\begin{bmatrix}
 \sigma_{xi}^2&\rho_i\sigma_{xi}\sigma_{yi}\\
 \rho_i\sigma_{xi}\sigma_{yi}&\sigma_{yi}^2
 \end{bmatrix},\\
 \sigma_{xi}>0,\qquad \sigma_{yi}>0, \qquad |\rho_i|<1,
 \label{covariance}
 \end{gathered}
\end{equation}
where $\sigma_{xi}$ and $\sigma_{yi}$ control spatial scales, while $\rho_i$ describes the correlation between the two spatial coordinates. At pixel position $\p$, let $\boldsymbol\delta_i=\p-\boldsymbol\mu_i^{xy}$ be the displacement from the center of primitive $i$.

We truncate each spatial Gaussian to limit its contribution to a local region. Using the squared distance $d_i^2(\p)=\boldsymbol\delta_i^\top(\Sigma_i^{xy})^{-1}\boldsymbol\delta_i$, the Gaussian weight becomes
\begin{equation}
 G_i(\p)=\exp\!\left[-\frac12
 \boldsymbol\delta_i^\top(\Sigma_i^{xy})^{-1}\boldsymbol\delta_i\right]
 \mathbf1[d_i^2(\p)\le\tau^2],
 \label{spatial_support}
\end{equation}
where $\tau$ controls the support. 

A spatial Gaussian primitive produces smoothly varying weights within a local region. To better represent fine spatial details, neural Gabor modulation is incorporated into the same local spatial primitive to adjust these weights at different positions. For primitive $i$, this modulation is defined as
\begin{equation}
 \begin{gathered}
 z_i(\p)=\sum_{h=1}^{D}\beta_{ih}\sin\left[\omega(\kappa^x_{ih}x_i+\kappa^y_{ih}y_i)+\phi_{ih}\right],
 \end{gathered}
 \label{gabor_preactivation}
\end{equation}
\begin{equation}
 \mathcal M_i(\p)=1+\eta\tanh z_i(\p),\qquad 
 0<\eta<1,
 \label{gabor}
\end{equation}
where $(x_i,y_i)$ are local spatial coordinates normalized by the covariance of primitive, $D$ is the number of sine functions, and $\omega$ controls their frequency scale. The frequencies $(\kappa^x_{ih},\kappa^y_{ih})$, phases $\phi_{ih}$, and weights $\beta_{ih}$ are predicted from local image features. The parameter $\eta$ controls the modulation strength and restricts $\mathcal M_i(\p)$ to $(1-\eta,1+\eta)$, keeping the primitive spatial weights nonnegative.  In this case, the primitive spatial weight is $\alpha_i(\p)=a_iG_i(\p)\mathcal M_i(\p)$. 

\subsubsection{Spatial-spectral coupling and Accumulation}
To retrieve spectral content from the dictionary, we predict mixture weights $\pi_{im}\ge0$ with $\sum_m\pi_{im}=1$. The dictionary mixture produces the primitive-specific spectrum
\begin{equation}
 d_i(\lambda)=\sum_m\pi_{im}s_m(\lambda),
 \label{joint}
\end{equation}
where $\pi_{im}$ controls the contribution of shared spectral primitive $m$ to local spatial primitive $i$.

We couple the local spatial primitive $\alpha_i(\p)$ with its dictionary-mixture spectrum $d_i(\lambda)$ to form a 3D spatial--spectral Gaussian primitive:
\begin{equation}
 \mathcal F_i(\p,\lambda) =a_iG_i(\p)\mathcal M_i(\p)\sum_m\pi_{im}s_m(\lambda).
 \label{modulated_primitive}
\end{equation}
After this coupling step, accumulation sums all coupled primitive contributions to construct the continuous spatial-spectral representation:
\begin{equation}
 \widehat{\mathcal H}(\p,\lambda)
 =\sum_i\mathcal F_i(\p,\lambda).
 \label{field}
\end{equation}

\subsection{Spectral Response Projection Rendering}
\label{rendering}
This section describes SRF integral projection and the subsequent spatial sampling and additive blending used to render the continuous representation into an RGB image.

\subsubsection{SRF integral projection}
For channel $c\in\{R,G,B\}$, the RGB observation is modeled by SRF integral projection of the continuous spatial–spectral representation. We apply this operation to each spatial-spectral primitive:
\begin{equation}
 g_{ic}(\p)=\int_0^1
 S_c(\lambda)\mathcal F_i(\p,\lambda)\,d\lambda.
 \label{project}
\end{equation}
For primitive $i$, the value $g_{ic}(\p)$ is its contribution to channel $c$ at spatial position $\p$. 

The primitive spatial weights in Eq.~\eqref{modulated_primitive} are wavelength-independent, so SRF integral projection can be performed on the spectral dictionary first:
\begin{equation}
 \begin{gathered}
 g_{ic}(\p)=a_iG_i(\p)\mathcal M_i(\p)\sum_m\pi_{im}
 \int_0^1 S_c(\lambda)s_m(\lambda)\,d\lambda,
 \end{gathered}
 \label{library_projection}
\end{equation}
where $r_{mc}=\int_0^1 S_c(\lambda)s_m(\lambda)\,d\lambda$ is the SRF integral projection of spectral primitive $m$ onto channel $c$.

For discrete computation at $J$ wavelengths $\{\lambda_j\}_{j=1}^{J}$, let $W_{cj}$ denote the SRF integration weight for channel $c$ at wavelength $\lambda_j$. Evaluating the spectral primitives at these wavelengths approximates the integral in Eq.~\eqref{library_projection}:
\begin{equation}
 r_{mc}\approx\sum_{j=1}^{J}W_{cj}s_m(\lambda_j),
 \label{discrete}
\end{equation}

Therefore, the contribution can be approximated as follows:
\begin{equation}
	g_{ic}(\p)
	\approx a_iG_i(\p)\mathcal M_i(\p)
	\sum_{m=1}^{M}\pi_{im}
	\sum_{j=1}^{J}W_{cj}s_m(\lambda_j).
	\label{discrete_projected_primitive}
\end{equation}

\subsubsection{Spatial sampling and additive blending}
After SRF integral projection, we sample the projected primitives at each pixel location $\p_{xy}=(x,y)$ and accumulate their RGB contributions. A pixel receives contributions from all primitives whose truncated spatial regions cover it, including those associated with neighboring anchors:

\begin{equation}
 \widehat Y_c(\p_{xy})
 =\sum_{a=1}^{HW}\sum_{k=1}^{K}g_{ak,c}(\p_{xy}),
 \label{blend}
\end{equation}
where $a$ indexes anchors in the double sum, $k$ indexes primitives within an anchor, and $x$ and $y$ index pixel columns and rows. The pair $(a,k)$ corresponds to the global primitive index $i=(a-1)K+k$, so $g_{ak,c}=g_{ic}$.

\subsection{Unsupervised loss}
\label{unsupervised_loss}

To constrain the rendered image to match the observation, we use the Charbonnier loss:
\begin{equation}
 \mathcal L_{\rm rgb}=\frac1{3HW}\sum_{\p,c}
 \sqrt{(\widehat Y_c(\p)-Y_c(\p))^2+\epsilon^2},
 \label{rgb_loss}
\end{equation}
where $\widehat Y$ is the rendered RGB image, $Y$ is the observed image, and $\epsilon>0$ smooths the absolute-error approximation near zero. 

To encourage spectral smoothness, we penalize the absolute second derivative of the reconstructed spectra with respect to wavelength:
\begin{equation}
 \mathcal L_{\rm spec}=\frac1{P_sJ_s}\sum_{r=1}^{P_s}\sum_{j=1}^{J_s}
 \left|\partial_{\lambda}^2\widehat{\mathcal H}(\p_r,\lambda_j^{\mathrm{reg}})\right|,
 \label{spectral_loss}
\end{equation}
where $P_s$ spatial points $\p_r$ are sampled uniformly within the image, and $J_s$ regularization wavelengths $\lambda_j^{\mathrm{reg}}$ are uniformly spaced on $[0,1]$, independently of the SRF integration nodes $\lambda_j$. The derivatives are evaluated analytically and their expressions are provided in the Supplementary Material (Sec.~A).

The weight $\gamma\ge0$ balances RGB rendering consistency and spectral regularization:
\begin{equation}
 \mathcal L=\mathcal L_{\rm rgb}+\gamma\mathcal L_{\rm spec}.
 \label{loss}
\end{equation}

\subsection{Inference}
\label{inference}
Given an input RGB image, the trained network estimates the spectral dictionary and spatial primitive parameters in a single forward pass. The continuous representation is established by coupling the spectral dictionary with the spatial primitives. For $L$ requested normalized wavelengths $\{\lambda_l\}_{l=1}^{L}\subset[0,1]$, spatial sampling and additive blending module directly produces the spectral band images without SRF integral projection:
\begin{equation}
	\widehat H_l[x,y]
	=\sum_{i=1}^{N}\alpha_i(\p_{xy})
	\sum_{m=1}^{M}\pi_{im}s_m(\lambda_l),
	\label{hsi_query}
\end{equation}
where $\alpha_i$ is the primitive spatial weight defined in Sec.~\ref{representation}, and $N=HWK$ is the total number of spatial primitives. Evaluating Eq.~\eqref{hsi_query} over the image grid and requested wavelengths yields the reconstructed HSI $\widehat H\in\R^{H\times W\times L}$.

\subsection{Optional post-hoc spectral refinement}
\label{sgre-method}

SRF integral projection imposes only three response-weighted integral constraints on each spectrum. Consequently, RGB rendering consistency alone cannot distinguish different spectra that produce identical sensor responses. To further constrain reconstruction, we combine full-spectrum adjustments with red-edge shape priors to propose an optional spectral refinement module. The module consists of two sequential steps: RGB-guided spectral refinement and fixed red-edge correction. This auxiliary module processes HSI data predicted by SpectralGS, without requiring the network to be retrained. The detailed formulation and implementation are provided in the Supplementary Material (Sec.~B).

\begin{table*}[t]
	\centering\footnotesize
	\setlength{\tabcolsep}{1.2pt}
	\renewcommand{\arraystretch}{1.10}
	\caption{Quantitative comparison on DFC2018, Chikusei, and Xiongan Matiwan. \tablebest{Best} and \tablesecond{second-best} values are highlighted.}
	\label{main}
	\begin{tabular*}{\textwidth}{@{\extracolsep{\fill}}l*{12}{r}@{}}
		\toprule
		&\multicolumn{4}{c}{DFC2018}&\multicolumn{4}{c}{Chikusei}&\multicolumn{4}{c}{Xiongan Matiwan}\\
		\cmidrule(lr){2-5}\cmidrule(lr){6-9}\cmidrule(lr){10-13}
		Method & PSNR$\uparrow$ & SSIM$\uparrow$ & SAM$\downarrow$ & RMSE$\downarrow$ & PSNR$\uparrow$ & SSIM$\uparrow$ & SAM$\downarrow$ & RMSE$\downarrow$ & PSNR$\uparrow$ & SSIM$\uparrow$ & SAM$\downarrow$ & RMSE$\downarrow$\\
		\midrule
		FSDFF & 18.8282 & 0.5714 & 29.9318 & 0.1553 & 12.9613 & 0.1077 & 56.7284 & 0.3014 & 8.2639 & 0.0119 & 70.7464 & 0.4697 \\
		SSRMamba & 23.6863 & 0.8147 & 18.0627 & 0.0913 & 24.7717 & 0.6369 & 26.0210 & 0.1148 & 13.7920 & 0.1446 & 42.1764 & 0.2652 \\
		RSNO & 25.6255 & 0.8873 & 14.2101 & 0.0690 & \tablesecond{28.2027} & \tablesecond{0.7777} & \tablesecond{15.8636} & 0.0871 & \tablesecond{26.4761} & \tablesecond{0.8500} & \tablebest{14.0897} & \tablebest{0.0751} \\
		DASDU & \tablesecond{26.1588} & \tablesecond{0.8937} & \tablesecond{13.3838} & \tablesecond{0.0651} & 24.9478 & 0.5603 & 28.6486 & 0.1227 & 25.7346 & 0.6556 & 17.0696 & 0.0882 \\
		MFormer & 22.0369 & 0.7217 & 22.9236 & 0.1086 & 17.2572 & 0.2782 & 37.3210 & 0.1840 & 18.9068 & 0.4971 & 30.0035 & 0.1625 \\
		UnSSR & 25.3072 & 0.8693 & 18.2311 & 0.0866 & 25.4933 & 0.7452 & 18.8620 & \tablesecond{0.0847} & 23.3969 & 0.8000 & 19.7251 & 0.1034 \\
		\textbf{SpectralGS (Ours)}& \tablebest{27.8392} & \tablebest{0.9090} & \tablebest{12.4008} & \tablebest{0.0589} & \tablebest{31.6980} & \tablebest{0.8901} & \tablebest{8.0666} & \tablebest{0.0412} & \tablebest{26.9405} & \tablebest{0.8667} & \tablesecond{14.3498} & \tablesecond{0.0773} \\
		\bottomrule
	\end{tabular*}
\end{table*}

\begin{figure*}[t]
	\centering
	\begingroup
	\setlength{\tabcolsep}{0.5pt}
	\renewcommand{\arraystretch}{0.2}
	\newcommand{\qimgw}{0.0735\textwidth}
	\newcommand{\qimg}[1]{\includegraphics[width=\qimgw]{figures/qualitative/assets/#1}}
	\begin{tabular}{@{}p{0.080\textwidth}*{4}{c}@{\hspace{0.008\textwidth}}*{4}{c}@{\hspace{0.008\textwidth}}*{4}{c}@{}}
		& \multicolumn{4}{c}{\small\textbf{DFC2018}} & \multicolumn{4}{c}{\small\textbf{Chikusei}} & \multicolumn{4}{c}{\small\textbf{Xiongan Matiwan}}\\[4pt]
		& \scriptsize SRF RGB & \scriptsize MAE & \scriptsize SRF RGB & \scriptsize MAE & \scriptsize SRF RGB & \scriptsize MAE & \scriptsize SRF RGB & \scriptsize MAE & \scriptsize SRF RGB & \scriptsize MAE & \scriptsize SRF RGB & \scriptsize MAE\\[3pt]
		\parbox[b][\qimgw][c]{0.078\textwidth}{\raggedleft\scriptsize MFormer\hspace{1.5pt}} &
		\qimg{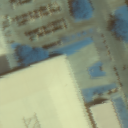} &
		\qimg{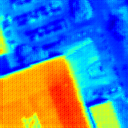} &
		\qimg{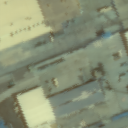} &
		\qimg{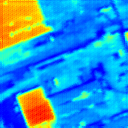} &
		\qimg{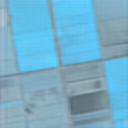} &
		\qimg{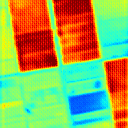} &
		\qimg{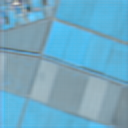} &
		\qimg{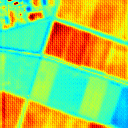} &
		\qimg{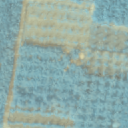} &
		\qimg{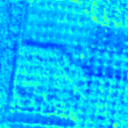} &
		\qimg{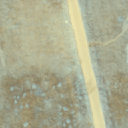} &
		\qimg{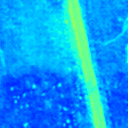}\\[1pt]
		\parbox[b][\qimgw][c]{0.078\textwidth}{\raggedleft\scriptsize UnSSR\hspace{1.5pt}} &
		\qimg{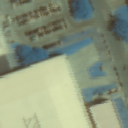} &
		\qimg{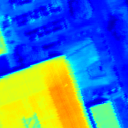} &
		\qimg{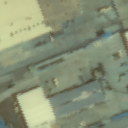} &
		\qimg{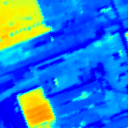} &
		\qimg{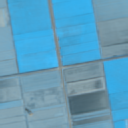} &
		\qimg{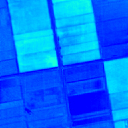} &
		\qimg{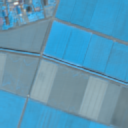} &
		\qimg{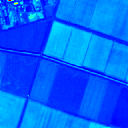} &
		\qimg{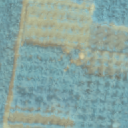} &
		\qimg{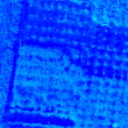} &
		\qimg{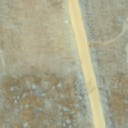} &
		\qimg{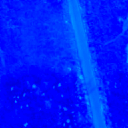}\\[1pt]
		\parbox[b][\qimgw][c]{0.078\textwidth}{\raggedleft\scriptsize SSRMamba\hspace{1.5pt}} &
		\qimg{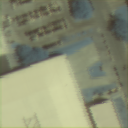} &
		\qimg{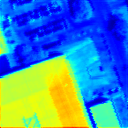} &
		\qimg{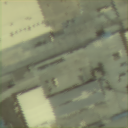} &
		\qimg{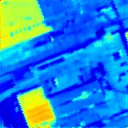} &
		\qimg{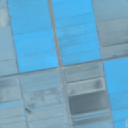} &
		\qimg{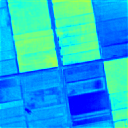} &
		\qimg{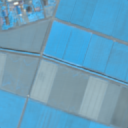} &
		\qimg{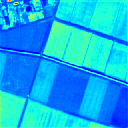} &
		\qimg{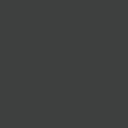} &
		\qimg{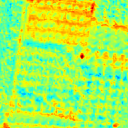} &
		\qimg{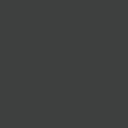} &
		\qimg{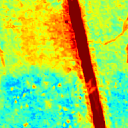}\\[1pt]
		\parbox[b][\qimgw][c]{0.078\textwidth}{\raggedleft\scriptsize RSNO\hspace{1.5pt}} &
		\qimg{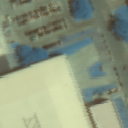} &
		\qimg{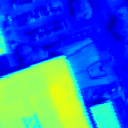} &
		\qimg{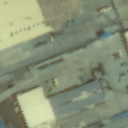} &
		\qimg{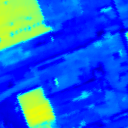} &
		\qimg{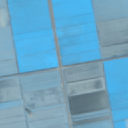} &
		\qimg{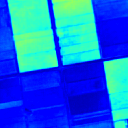} &
		\qimg{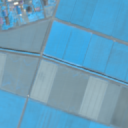} &
		\qimg{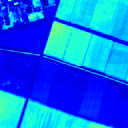} &
		\qimg{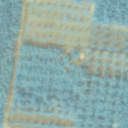} &
		\qimg{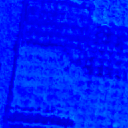} &
		\qimg{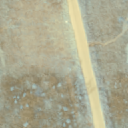} &
		\qimg{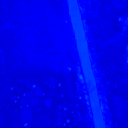}\\[1pt]
		\parbox[b][\qimgw][c]{0.078\textwidth}{\raggedleft\scriptsize FSDFF\hspace{1.5pt}} &
		\qimg{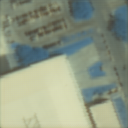} &
		\qimg{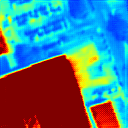} &
		\qimg{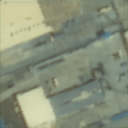} &
		\qimg{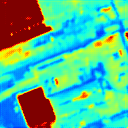} &
		\qimg{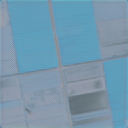} &
		\qimg{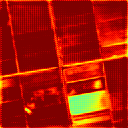} &
		\qimg{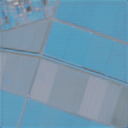} &
		\qimg{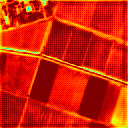} &
		\qimg{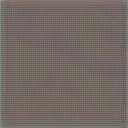} &
		\qimg{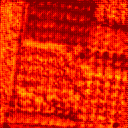} &
		\qimg{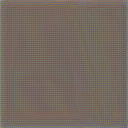} &
		\qimg{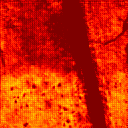}\\[1pt]
		\parbox[b][\qimgw][c]{0.078\textwidth}{\raggedleft\scriptsize DASDU\hspace{1.5pt}} &
		\qimg{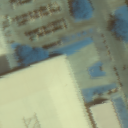} &
		\qimg{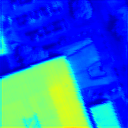} &
		\qimg{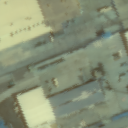} &
		\qimg{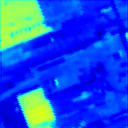} &
		\qimg{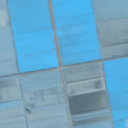} &
		\qimg{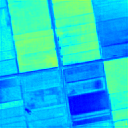} &
		\qimg{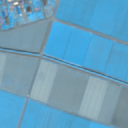} &
		\qimg{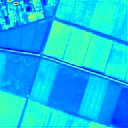} &
		\qimg{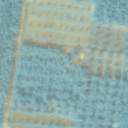} &
		\qimg{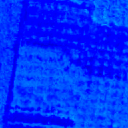} &
		\qimg{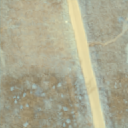} &
		\qimg{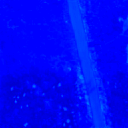}\\[1pt]
		\parbox[b][\qimgw][c]{0.078\textwidth}{\raggedleft\scriptsize \textbf{SpectralGS}\hspace{1.5pt}} &
		\qimg{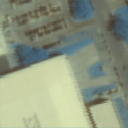} &
		\qimg{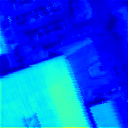} &
		\qimg{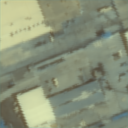} &
		\qimg{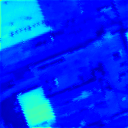} &
		\qimg{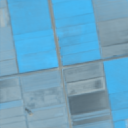} &
		\qimg{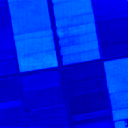} &
		\qimg{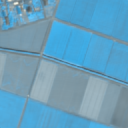} &
		\qimg{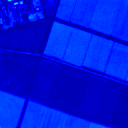} &
		\qimg{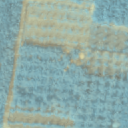} &
		\qimg{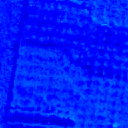} &
		\qimg{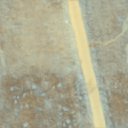} &
		\qimg{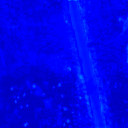}\\[1pt]
		\parbox[b][\qimgw][c]{0.078\textwidth}{\raggedleft\scriptsize GT\hspace{1.5pt}} &
		\qimg{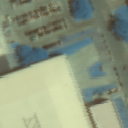} &
		\qimg{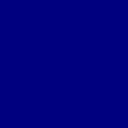} &
		\qimg{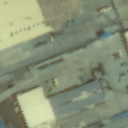} &
		\qimg{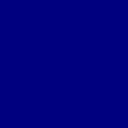} &
		\qimg{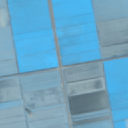} &
		\qimg{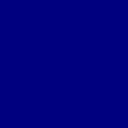} &
		\qimg{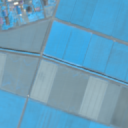} &
		\qimg{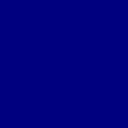} &
		\qimg{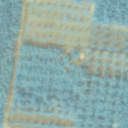} &
		\qimg{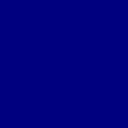} &
		\qimg{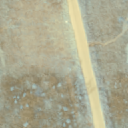} &
		\qimg{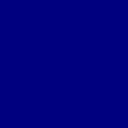}\\[1pt]
		\scriptsize HSI MAE & \multicolumn{4}{c}{\includegraphics[width=0.27\textwidth]{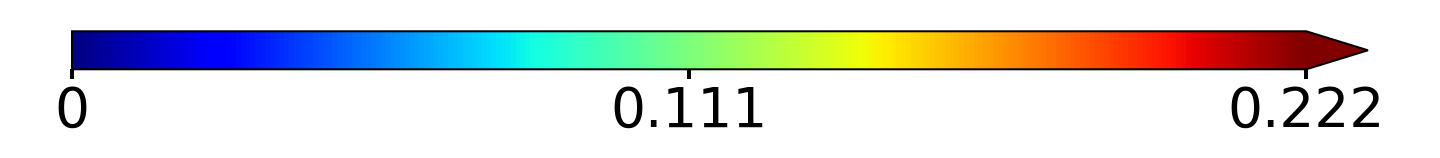}} & \multicolumn{4}{c}{\includegraphics[width=0.27\textwidth]{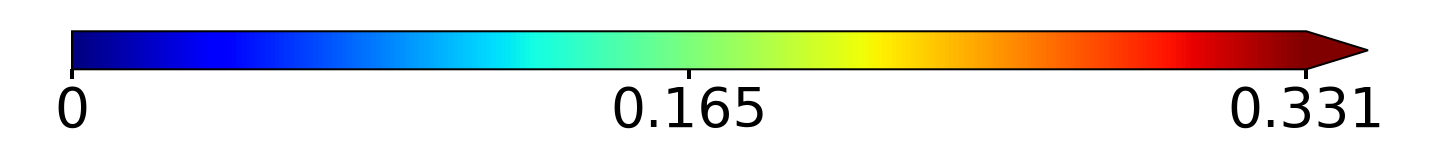}} & \multicolumn{4}{c}{\includegraphics[width=0.27\textwidth]{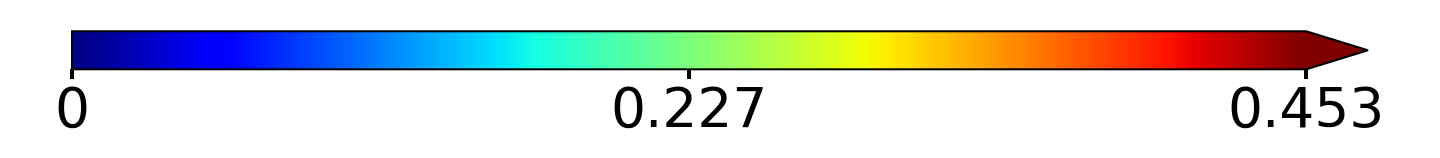}}\\
	\end{tabular}
	\endgroup
	\caption{Qualitative comparisons on DFC2018, Chikusei, and Xiongan Matiwan. Each sample is shown as an SRF-projected three-channel image (left) and an HSI MAE map averaged over all spectral bands (right). }
	\label{qualitative}
\end{figure*}

\section{Experiments}
We evaluate SpectralGS on three hyperspectral datasets: DFC2018 Houston dataset \footnote[1]{\href{https://hyperspectral.ee.uh.edu/?page_id=1075}{\text{https://hyperspectral.ee.uh.edu/?page\_id=1075}}}, Chikusei dataset \cite{2016Airborne}, and Xiongan Matiwan dataset \cite{cen2020aerial}. This section presents quantitative and qualitative comparisons with existing methods, followed by quantitative and qualitative ablation studies to assess individual components, parameter studies to examine the effect of the spectral dictionary size, and an evaluation of post-hoc spectral refinement.

\subsection{Compared algorithms and experimental settings}
\label{compared_methods}

\textbf{Compared algorithms.}
Six baseline methods are included in the comparison: FSDFF~\cite{10494781}, SSRMamba~\cite{10887701}, RSNO~\cite{zhang2026radiativestructuredneuraloperatorcontinuous}, DASDU~\cite{WANG2026114641}, MFormer~\cite{10093998}, and UnSSR~\cite{10843147}. MFormer and UnSSR retain their original unsupervised objectives. The other four supervised algorithms are adapted to RGB-only training with RGB rendering consistency and spectral regularization.

\noindent
\textbf{Data preparation and SRFs.}
We extract overlapping $128\times128$ patches from each hyperspectral dataset to construct the training, validation, and test sets. Synthetic SRF is used in the experiments on the DFC2018 dataset. For Chikusei and Xiongan Matiwan, the supplied SRFs are based on the SRF of the ELECTRO-L 2 MSUGS sensor. \footnote[2]{\href{https://nwp-saf.eumetsat.int/downloads/rtcoef_info/visir_srf/rtcoef_electro-l_2_msugs_srf.html}{\text{https://nwp-saf.eumetsat.int/}\allowbreak\text{downloads/rtcoef\_info/visir\_srf/}\allowbreak\text{rtcoef\_electro-l\_2\_msugs\_srf.html}}}

\noindent
\textbf{Implementation details.}
Each anchor has $K=8$ spatial primitives. The dictionary has $M=8$ spectral primitives with $Q=8$ Gaussian kernels each. Gabor parameters are $D=8$, $\omega=6$, and $\eta=0.5$, with spatial truncation radius $\tau=3$. Loss parameters are $\epsilon=10^{-3}$ and $\gamma=3\times10^{-5}$.

Throughout the main comparison, SpectralGS denotes the two-component framework without post-hoc refinement. The component ablations and dictionary-size study use the same inference procedure.

\begin{figure}[t]
	\centering
	\includegraphics[width=0.9\columnwidth]{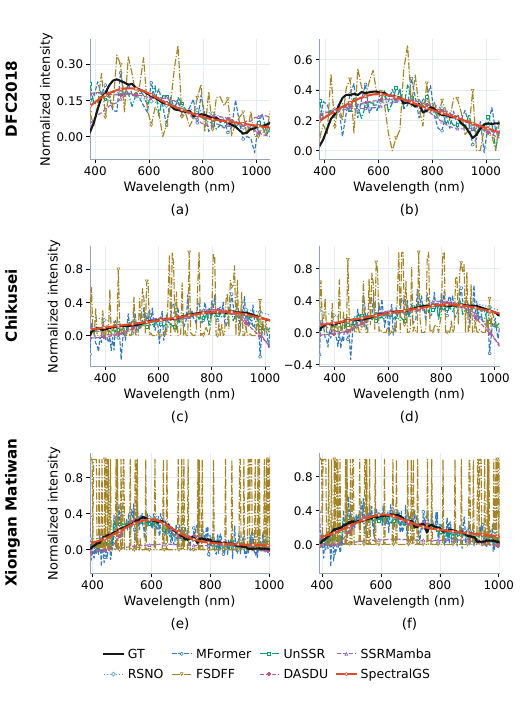}
	\caption{Pixel-level spectral comparisons on DFC2018, Chikusei, and Xiongan Matiwan (top to bottom).}
	\label{spectra}
\end{figure}

\subsection{Quantitative and qualitative comparison}
\label{quantitative}
Table~\ref{main} presents the quantitative results, complemented by the SRF-projected RGB images and MAE maps in Fig.~\ref{qualitative} and the spectral curves in Fig.~\ref{spectra}. 

\noindent
\textbf{DFC2018.}
SpectralGS outperforms the compared methods on all four reported metrics. Its MAE maps show that errors are smaller within large rectangular regions and that there are fewer areas where errors are concentrated. The spectral curves selected in Fig.~\ref{spectra} (a,b) generally follow the intensity variation trend of the ground truth (GT) and exhibit fewer irregular fluctuations compared to MFormer and FSDFF. Although certain severe spectral variations are still smoothed out, the reconstructed spectral curves generally show a higher degree of consistency with the GT.

\noindent
\textbf{Chikusei.}
SpectralGS shows the greatest improvement on Chikusei, with gains in both spatial reconstruction and spectral accuracy. The MAE maps consistently show lower error values across the displayed regions. In Fig.~\ref{spectra} (c,d), the reconstructed spectra closely follow the gradually rising, broad-peaked and subsequently declining GT curve. Compared with other methods, SpectralGS better preserves the shape and intensity of the spectra, particularly at the long-wavelength end.

\noindent
\textbf{Xiongan Matiwan.}
SpectralGS obtains the highest PSNR and SSIM on Xiongan Matiwan, while RSNO achieves lower SAM and RMSE. Their MAE maps exhibit similar error patterns, with both SpectralGS and RSNO exhibiting significantly better performance than the other comparison methods. At the selected pixels in Fig.~\ref{spectra}(e,f), SpectralGS more closely reproduces the broad spectral peaks and their decline towards longer wavelengths, with fewer distortions than several competing methods. SSRMamba and FSDFF exhibit greater intensity deviations or spectral fluctuations in these examples.

\subsection{Ablation studies}
\label{diagnostics}
We conduct ablations on the Chikusei dataset by replacing the shared dictionary with an independent spectrum for each spatial primitive, removing Gabor modulation, or removing spectral regularization. Each variant is trained separately with the remaining settings unchanged. 

\begin{table}[t]
\centering
\footnotesize
\setlength{\tabcolsep}{2pt}
\renewcommand{\arraystretch}{1.15}
\caption{Component ablations on Chikusei over all 128 bands. \tablebest{Best} and \tablesecond{second-best} values are highlighted.}
\label{ablation}
\begin{tabular*}{\columnwidth}{@{\extracolsep{\fill}}lcccc@{}}
\toprule
Variant & PSNR$\uparrow$ & SSIM$\uparrow$ & SAM$\downarrow$ & RMSE$\downarrow$\\
\midrule
SpectralGS & \tablebest{31.6980} & \tablebest{0.8901} & \tablebest{8.0666} & \tablebest{0.0412}\\
Independent spectra & 21.3843 & 0.5723 & 30.9876 & 0.1176\\
w/o Gabor & \tablesecond{29.4442} & \tablesecond{0.8772} & \tablesecond{10.7482} & \tablesecond{0.0559}\\
w/o spectral reg. & 21.3453 & 0.6937 & 62.0707 & 0.8208\\
\bottomrule
\end{tabular*}
\end{table}

\begin{figure}[t]
	\centering
	\includegraphics[width=\columnwidth]{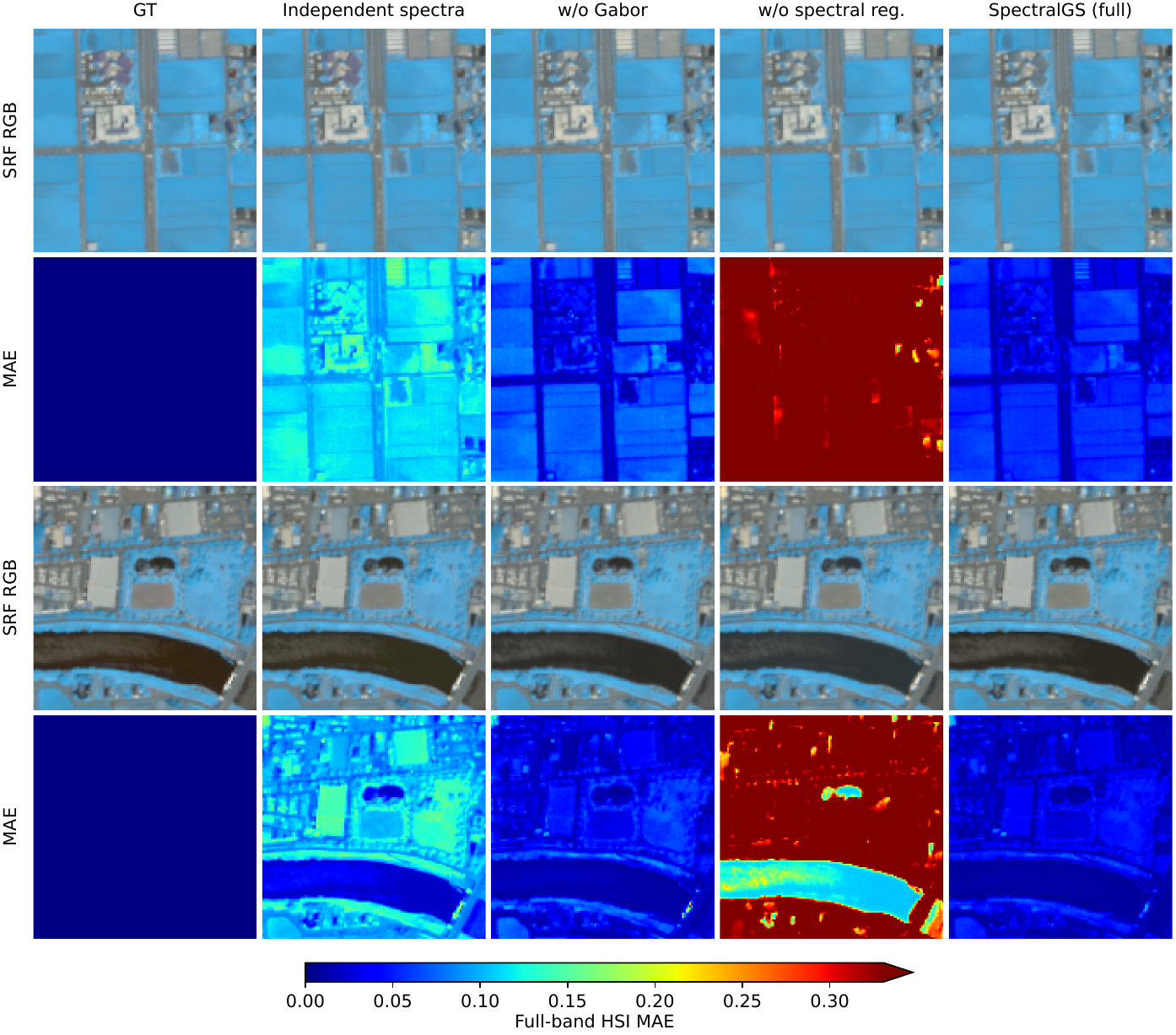}
	\caption{Qualitative ablations on Chikusei.}
	\label{ablation_qualitative}
\end{figure}

SpectralGS performs best across all metrics in Table~\ref{ablation}. Using independent spectra reduces PSNR by 10.31\,dB, while removing Gabor modulation causes a smaller drop of 2.25\,dB. Removing spectral regularization yields the largest RMSE and SAM, highlighting its role in constraining spectral recovery. Fig.~\ref{ablation_qualitative} shows corresponding increases in spatial errors, particularly for independent spectra and the variant without spectral regularization. Their broadly similar RGB appearances indicate that RGB rendering consistency alone does not ensure accurate HSI reconstruction.

\subsection{Effect of spectral dictionary size}
\label{dictionary_size_study}
We vary $M\in\{2,4,8,16,32\}$ on DFC2018 with $K=Q=8$. Full numerical results and dictionary visualizations are in the Supplementary Material (Sec.~C).

Fig.~\ref{dictionary_metrics} shows the non-monotonic trend of the metrics as the dictionary size varies. When $M=32$, RMSE reaches its lowest value, while PSNR and SSIM reach their highest values. In contrast, when $M=4$, SAM exhibits the lowest value. Compared to $M=8$, $M=32$ improves PSNR and SSIM, but increases SAM. However, compared to $M=2$, $M=32$ shows improvements across all four metrics. Therefore, a larger dictionary can improve reconstruction performance, but the specific gains and the optimal dictionary size depend on the evaluation metrics.

\begin{figure}[t]
	\centering
	\includegraphics[width=0.85\columnwidth]{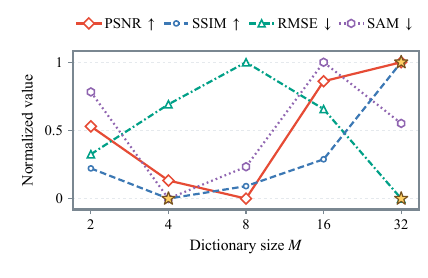}
	\caption{Reconstruction metrics versus spectral dictionary size on DFC2018, independently min--max scaled to $[0,1]$. Arrows indicate preferred directions. Stars mark metric-specific optima: PSNR, SSIM, and RMSE at $M=32$, and SAM at $M=4$. Original values are in the Supplementary Material.}
	\label{dictionary_metrics}
\end{figure}

\subsection{Evaluation of optional post-hoc spectral refinement}
\label{sgre-results}
Table~\ref{sgre-ablation} summarizes RGB-guided spectral refinement followed by fixed red-edge correction applied to fixed SpectralGS predictions.

\begin{table}[htbp]
\centering
\caption{Sequential refinement. \tablebest{Best} and \tablesecond{second-best} values are highlighted per dataset, including displayed ties.}
\label{sgre-ablation}
\begingroup\footnotesize
\setlength{\tabcolsep}{1.5pt}
\renewcommand{\arraystretch}{1.08}
\begin{tabular*}{\columnwidth}{@{\extracolsep{\fill}}llrrrr@{}}
\toprule
Dataset & Output & PSNR$\uparrow$ & SSIM$\uparrow$ & SAM$\downarrow$ & RMSE$\downarrow$\\
\midrule
\multirow{3}{*}{DFC2018}
 & Unrefined & 27.8392 & 0.9090 & \tablesecond{12.4008} & \tablesecond{0.0589}\\
 & + RGB-guided & \tablesecond{28.2906} & \tablesecond{0.9150} & 12.4449 & 0.0594\\
 & + Red-edge & \tablebest{29.0773} & \tablebest{0.9238} & \tablebest{11.9754} & \tablebest{0.0569}\\
\midrule
\multirow{3}{*}{Chikusei}
 & Unrefined & 31.6980 & 0.8901 & 8.0666 & \tablesecond{0.0412}\\
 & + RGB-guided & \tablesecond{32.3478} & \tablesecond{0.8937} & \tablesecond{7.9455} & \tablesecond{0.0412}\\
 & + Red-edge & \tablebest{34.1820} & \tablebest{0.9130} & \tablebest{6.7307} & \tablebest{0.0324}\\
\midrule
\multirow{3}{*}{\shortstack[l]{Xiongan\\Matiwan}}
 & Unrefined & 26.9405 & 0.8667 & 14.3498 & 0.0773\\
 & + RGB-guided & \tablesecond{27.0224} & \tablesecond{0.8701} & \tablesecond{14.2290} & \tablesecond{0.0766}\\
 & + Red-edge & \tablebest{27.3414} & \tablebest{0.8771} & \tablebest{13.7412} & \tablebest{0.0737}\\
\bottomrule
\end{tabular*}
\endgroup
\end{table}

RGB-guided spectral refinement improves PSNR and SSIM on all datasets, but slightly worsens SAM and RMSE on DFC2018. The subsequent red-edge correction improves all four metrics on every dataset. Relative to the unrefined predictions, the combined steps increase PSNR on DFC2018, Chikusei, and Xiongan Matiwan, respectively. Chikusei shows the largest PSNR gain, with SAM also decreasing. These results support the shape prior as a complement to RGB-guided fitting. Additional visual comparisons are provided in the Supplementary Material (Sec.~B).

\FloatBarrier

\section{Conclusion}
We propose SpectralGS, a Gaussian splatting framework for unsupervised spectral super-resolution, which integrates a structured spatial–spectral representation with sensor image processing. Shared Spectral Gaussian Representation employs a spectral dictionary with local spatial Gaussian primitives to incorporate a low-rank spectral prior. Spectral Response Projection Rendering applies SRF integral projection, spatial sampling and additive blending to generate RGB images, enabling training through RGB rendering consistency. Experiments on three datasets demonstrate the reconstruction performance of SpectralGS.

\clearpage
\section*{LLM Usage Statement}
We used ChatGPT for language polishing and grammatical correction. All technical content and conclusions were developed and verified solely by the authors.

\clearpage
\setcounter{page}{1}
\maketitlesupplementary
\appendix
\raggedbottom
\section{Analytic derivatives for spectral regularization}
\label{spectral_derivatives}
The spectral regularization in Eq.~\eqref{spectral_loss} penalizes the absolute second derivative of the complete continuous representation with respect to normalized wavelength $\lambda$. For the Gaussian kernels in Eq.~\eqref{spectral_kernel},
\begin{equation}
 b_{mq}''(\lambda)=b_{mq}(\lambda)
 \left[\frac{(\lambda-\nu_{mq})^2}{w_{mq}^4}-\frac1{w_{mq}^2}\right].
 \label{kernel_curvature}
\end{equation}
The normalization $Z_m$ in Eq.~\eqref{library} is independent of the query wavelength, giving
\begin{equation}
 s_m''(\lambda)=\frac1{Z_m}\sum_q c_{mq}b_{mq}(\lambda)
 \left[\frac{(\lambda-\nu_{mq})^2}{w_{mq}^4}-\frac1{w_{mq}^2}\right].
 \label{library_curvature}
\end{equation}
Since the primitive spatial weights and mixture weights are wavelength-independent, the derivative of the continuous representation is
\begin{equation}
 \partial_{\lambda}^2\widehat{\mathcal H}(\p,\lambda)
 =\sum_i a_iG_i(\p)\mathcal M_i(\p)\sum_m\pi_{im}s_m''(\lambda).
 \label{curvature}
\end{equation}
We evaluate this expression at $(\p_r,\lambda_j^{\mathrm{reg}})$, take the absolute value after summing all contributions, and average as in Eq.~\eqref{spectral_loss}. Although $Z_m$ is constant with respect to the query wavelength, it remains differentiable for the spectral dictionary parameters during training.

\section{Optional post-hoc spectral refinement}
\label{refinement}
SRF integral projection imposes only three response-weighted integral constraints on each spectrum. Consequently, RGB rendering consistency alone cannot distinguish different spectra that produce identical sensor responses. To further constrain reconstruction, we combine full-spectrum adjustment with a spectral shape prior. 

First, we use RGB observations to adjust the shared spectral dictionary by optimizing the centers, widths, and composition coefficients of its Gaussian kernels, with RGB rendering consistency and nonnegativity enforced on the output. We then introduce a red-edge shape prior. Specifically, vegetation often covers large areas in remote sensing imagery, and its reflectance typically rises sharply between \(680\!-\!750\,\mathrm{nm}\), forming the red edge~\cite{sgreHorler1983}. This rapid transition is difficult to recover from RGB alone and may be weakened by spectral smoothing. We therefore use a fixed red-edge template to guide its reconstruction.

\subsection{Refinement formulation}
\label{refinement_formulation}
\paragraph{RGB-guided spectral refinement.}
We adjust the spectral dictionary parameters $\Theta_{\mathrm{spec}}$ and a shared offset to the mixture-weight logits through $\mathbf z$, with $\mathbf z=0$ denoting the initial prediction. The predictor weights $\theta$ and primitive spatial weights $\alpha_i(\p)$ remain fixed. Using the rendered RGB $\widehat Y_c(\p,\mathbf z)$, we solve
\begin{equation}
 \widehat{\mathbf z}=\arg\min_{\mathbf z}
 \frac{\sum_{\p,c}|\widehat Y_c(\p,\mathbf z)-Y_c(\p)|^2}{2\sigma_{\mathrm{rgb}}^2}
 +\frac{\|\mathbf z\|_2^2}{2\tau_z^2}.
 \label{sgre-iid}
\end{equation}
The two terms balance RGB fitting and parameter changes through the positive scales $\sigma_{\mathrm{rgb}}$ and $\tau_z$. This corresponds to a maximum a posteriori estimate with independent, equal-variance Gaussian residuals and a zero-centered Gaussian prior on $\mathbf z$.

Let $\widehat{\mathbf H}(\p,\mathbf z)=[\widehat{\mathcal H}_{\Theta(\mathbf z)}(\p,\lambda_l)]_{l=1}^{L}\in\mathbb R^L$ denote the spectrum at pixel $\p$. The SRF weight matrix $A=[W_{cl}]\in\mathbb R^{3\times L}$ therefore maps this spectrum to the rendered RGB vector, $\widehat Y(\p,\mathbf z)=A\widehat{\mathbf H}(\p,\mathbf z)$. Since fitting does not enforce exact RGB rendering consistency, we find the closest nonnegative spectrum that satisfies the observation:
\begin{equation}
 \mathbf b(\p)=\arg\min_{\substack{\mathbf h\geq0\\A\mathbf h=Y(\p)}}
 \|\mathbf h-\widehat{\mathbf H}(\p,0)\|_2^2.
 \label{sgre-anchor}
\end{equation}
We retain the fitted spectral change without altering RGB by projecting it with $P=I-A^\dagger A$, where $A^\dagger$ is the pseudoinverse and $AP=0$:
\begin{equation}
 \begin{aligned}
 \widehat{\mathbf H}^{\mathrm I}(\p)={}&\mathbf b(\p)+t_{\p}P\big[
 \widehat{\mathbf H}(\p,\widehat{\mathbf z})
 -\widehat{\mathbf H}(\p,0)\big].
 \end{aligned}
 \label{sgre-iid-output}
\end{equation}
The step $t_{\p}$ is the largest value in $[0,1]$ that keeps all bands nonnegative. The output combines an RGB-consistent reference with the fitted change in the null space of $A$.

\begin{figure*}[t]
	\centering
	\includegraphics[width=\textwidth,trim=28.45bp 0bp 0bp 0bp,clip]{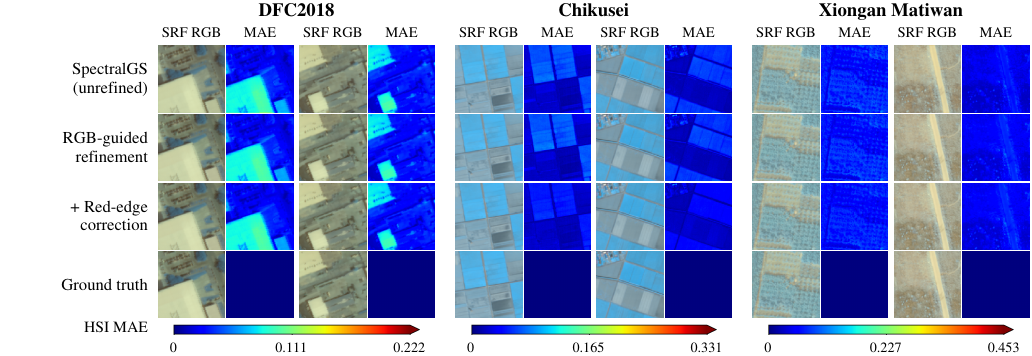}
	\caption{SRF-projected images and full-band MAE maps before and after the two sequential refinement steps, with GT for reference.}
	\label{sgre-images}
\end{figure*}

\paragraph{Fixed red-edge correction.}
We use a fixed vegetation template to guide spectral shape beyond RGB rendering consistency. For physical wavelength $\lambda^{\mathrm{nm}}$, the template is inspired by inverted-Gaussian vegetation models~\cite{sgreMiller1990}:
\begin{equation}
 \begin{aligned}
 E(\lambda^{\mathrm{nm}})={}&1-\exp\!\left[-\frac{[\lambda^{\mathrm{nm}}-675]_+^2}{2(45)^2}\right]\\
 &+0.15\exp\!\left[-\frac{(\lambda^{\mathrm{nm}}-550)^2}{2(25)^2}\right],
 \end{aligned}
 \label{sgre-template}
\end{equation}
Here, $[x]_+=\max(x,0)$. The two terms describe the red-edge rise and a smaller green-band shoulder. We sample the template at the $L$ output wavelengths $\lambda_l^{\mathrm{nm}}$ to obtain $\mathbf E=[E(\lambda_l^{\mathrm{nm}})]_{l=1}^{L}\in\mathbb R^L$.

The SRF weight $A_{cl}$ specifies the contribution of wavelength $l$ to channel $c$. Its response centroid, $\bar\lambda_c^{\mathrm{nm}}=\sum_l A_{cl}\lambda_l^{\mathrm{nm}}$, is the response-weighted mean wavelength of that channel. We select the channels with the lowest and highest centroids as $c_-$ and $c_+$, respectively. Their contrast is used below to scale and activate the correction.

The template contains both broad spectral variation and a rapid red-edge rise. We subtract a smoothed copy to suppress its slowly varying components and emphasize local spectral changes that may be weakened by spectral regularization. The smoothed template is $\mathbf G\mathbf E$, where $\mathbf G\in\mathbb R^{L\times L}$ averages template samples along the wavelength dimension and has entries
\begin{equation}
 G_{lk}=\frac{\exp\!\left[-\frac{(\lambda_l^{\mathrm{nm}}-\lambda_k^{\mathrm{nm}})^2}{2\sigma_G^2}\right]}{\sum_{j=1}^{L}\exp\!\left[-\frac{(\lambda_l^{\mathrm{nm}}-\lambda_j^{\mathrm{nm}})^2}{2\sigma_G^2}\right]}.
 \label{sgre-smoothing}
\end{equation}
Thus, $(\mathbf G\mathbf E)_l$ averages nearby template samples, with the bandwidth $\sigma_G$ controlling the wavelength range of the averaging. We set $\sigma_G=(\bar\lambda_{c_+}^{\mathrm{nm}}-\bar\lambda_{c_-}^{\mathrm{nm}})/4$ as a fixed design choice tied to the separation of the selected SRF channels. Around a rising transition, the residual $\mathbf E-\mathbf G\mathbf E$ can have negative values before the rise and positive values after it, providing a direction for increasing local spectral contrast. We then project this residual into the null space of $A$ to preserve RGB rendering consistency:
\begin{equation}
 \mathbf d=\frac{P(\mathbf E-\mathbf G\mathbf E)}{(A_{c_+,:}-A_{c_-,:})\mathbf E}.
 \label{sgre-direction}
\end{equation}
Here, \(A_{c_+,:}\) and \(A_{c_-,:}\) denote the rows of the SRF matrix corresponding to the channels with the highest and lowest response centroids, respectively.
The positive denominator scales the direction by outer-channel contrast of the template. The pixelwise strength is $\zeta(\p)=\tfrac12[Y_{c_+}(\p)-Y_{c_-}(\p)]_+$. The final spectrum is
\begin{equation}
 \widehat{\mathbf H}^{\mathrm F}(\p)=\widehat{\mathbf H}^{\mathrm I}(\p)+t_{\p}^{\mathrm{re}}\zeta(\p)\mathbf d,
 \label{sgre-final}
\end{equation}
The step $t_{\p}^{\mathrm{re}}$ is the largest value in $[0,1]$ preserving nonnegativity. Since $A\mathbf d=0$, the update preserves the RGB rendering consistency established in Eq.~\eqref{sgre-anchor}. Smoothing and null-space projection can also distribute the correction beyond the red-edge interval.

\subsection{Evaluation results}

We evaluate the two cumulative refinement steps on DFC2018, Chikusei, and Xiongan Matiwan patches using the RGB observations, known SRF, and fixed network predictions. 

Fig.~\ref{sgre-images} compares the SRF-projected images and full-band MAE maps before and after the two refinement steps. RGB-guided spectral refinement produces relatively modest changes in the error maps, while the subsequent red-edge correction provides more visible improvements on Chikusei, where errors decrease across several rectangular field regions. These observations complement the quantitative results in Table~\ref{sgre-ablation} of the main paper.

The SRF-projected images remain visually unchanged between the two refinement stages, although their HSI error maps differ. This illustrates that red-edge correction can adjust the reconstructed spectra while preserving RGB rendering consistency.

\section{Spectral dictionaries and spatial mixture fractions}
\label{dictionary_visualizations}

\begin{figure*}[t]
\centering
\includegraphics[width=\textwidth]{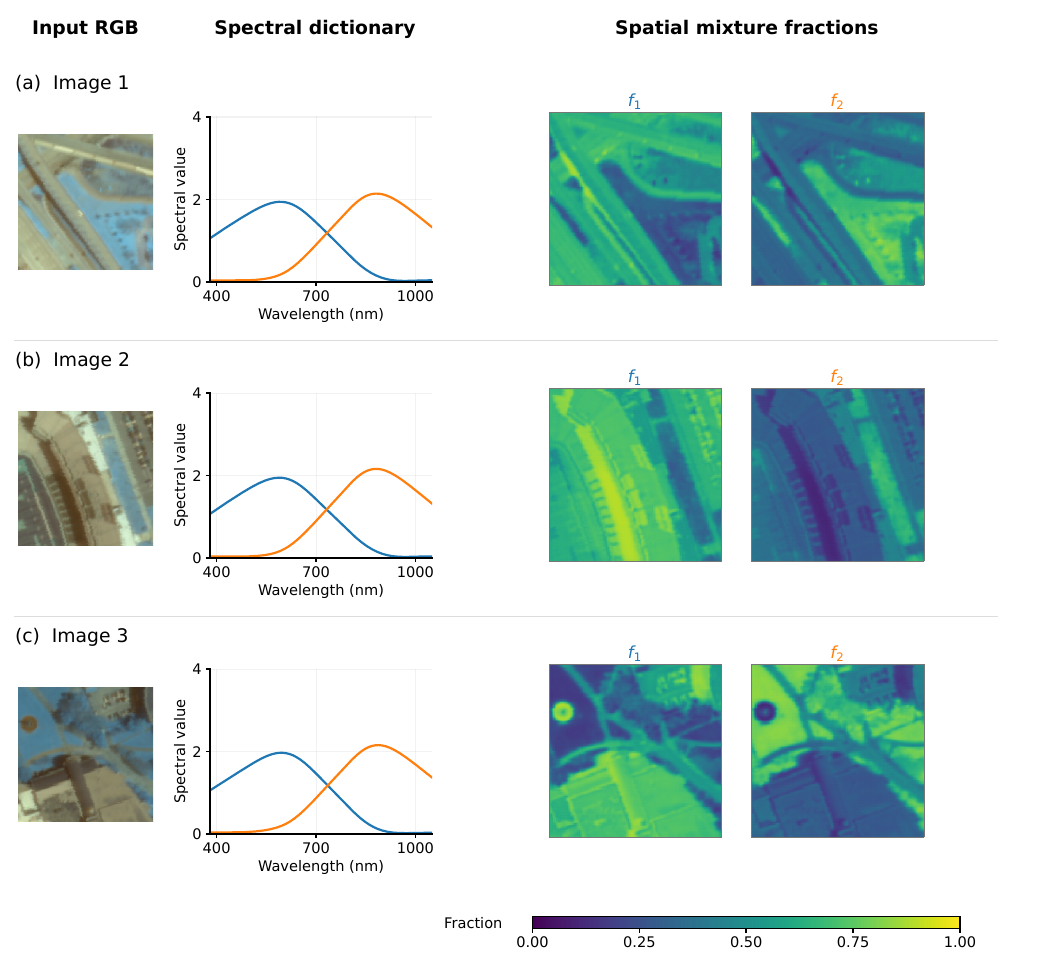}
\caption{Spectral dictionary and spatial mixture fractions with $M=2$ on three DFC2018 images.}
\label{dictionary_size_examples}
\end{figure*}

\begin{figure*}[t]
\centering
\includegraphics[width=\textwidth]{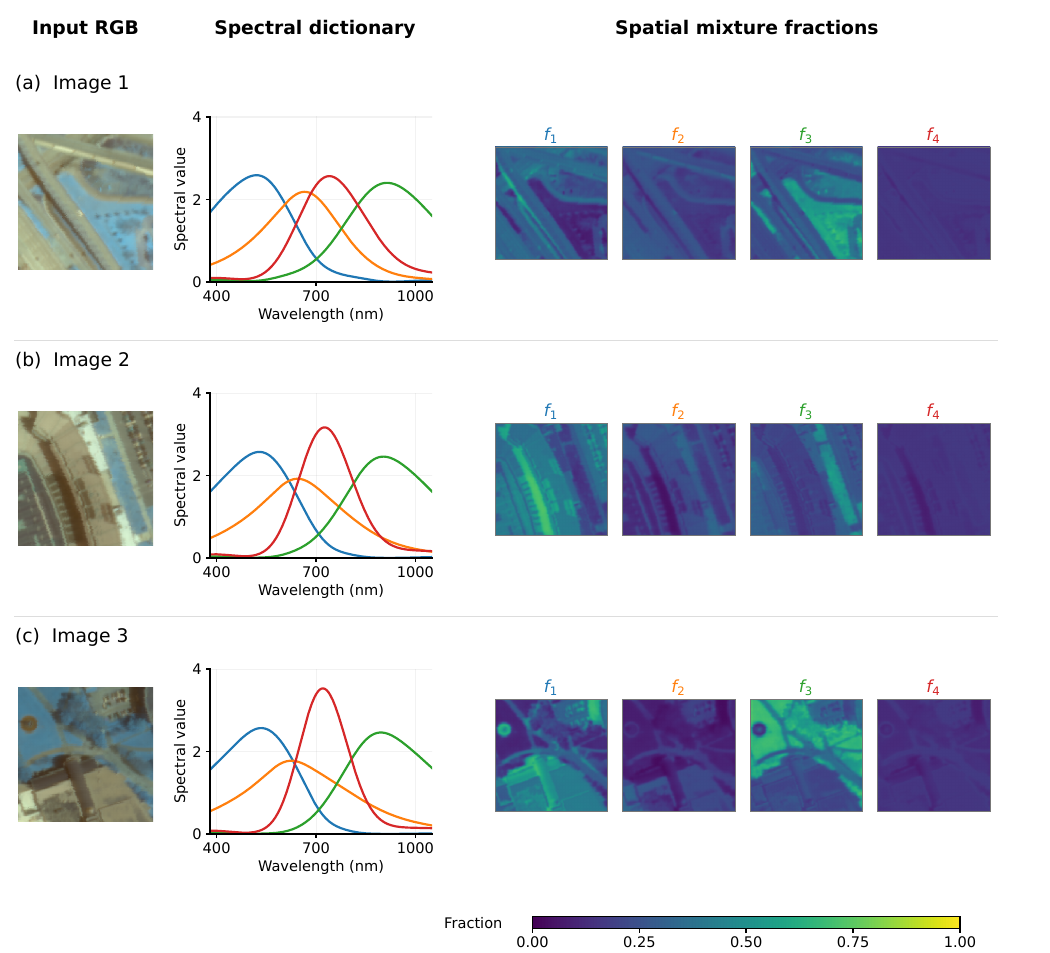}
\caption{Spectral dictionary and spatial mixture fractions with $M=4$ on the same three DFC2018 images.}
\label{dictionary_size_m4}
\end{figure*}

\begin{figure*}[t]
\centering
\includegraphics[width=\textwidth]{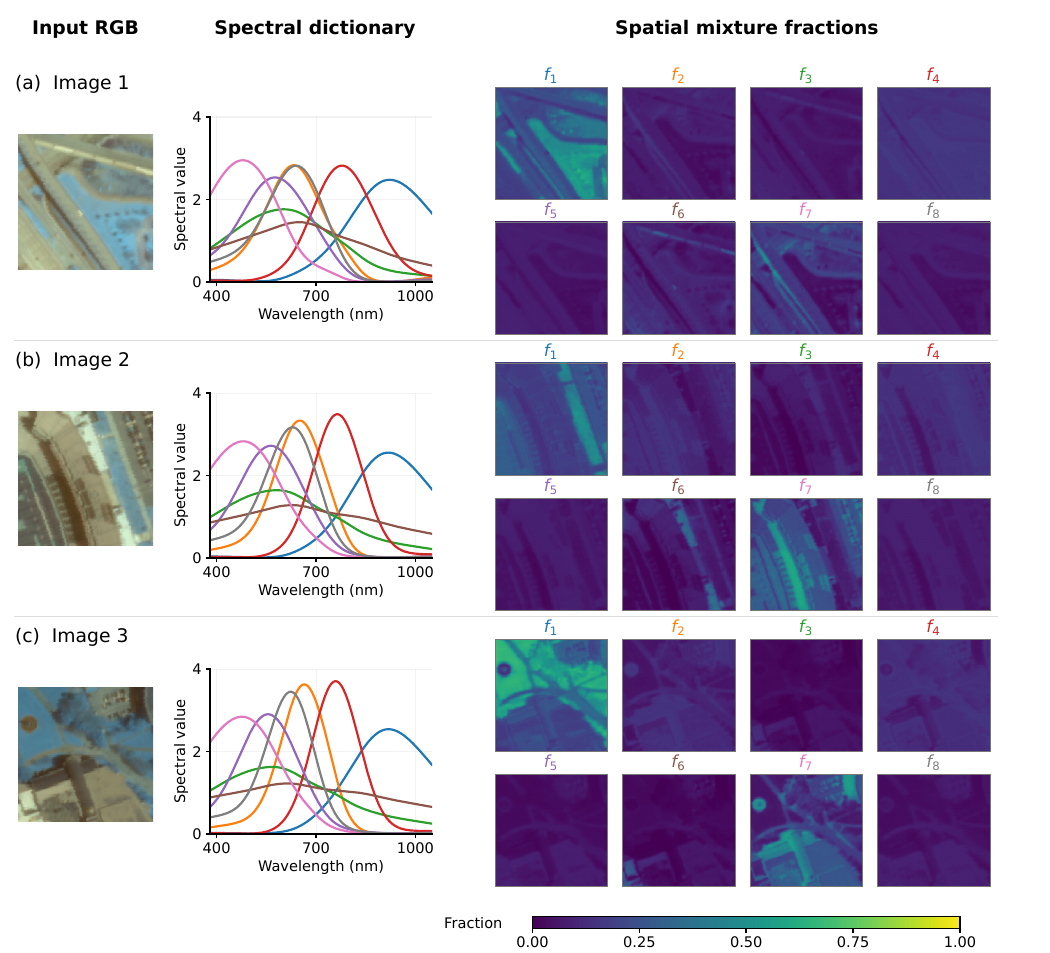}
\caption{Spectral dictionary and spatial mixture fractions with $M=8$ on the same three DFC2018 images. }
\label{dictionary_size_m8}
\end{figure*}

\begin{figure*}[p]
\centering
\includegraphics[width=\textwidth]{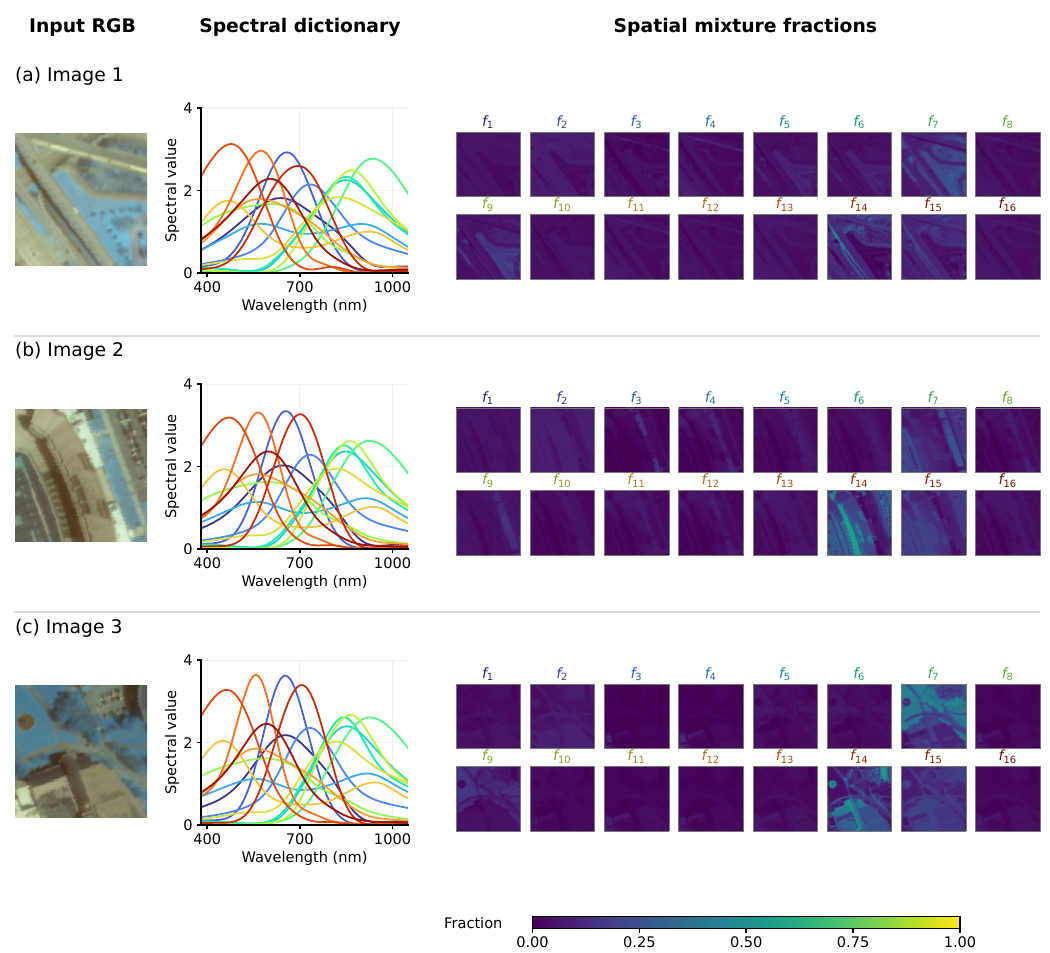}
\caption{Spectral dictionary and spatial mixture fractions with $M=16$ on the same three DFC2018 images. All 16 spectral primitives and their fraction maps are shown. Fraction maps use the common $[0,1]$ scale.}
\label{dictionary_size_m16}
\end{figure*}

\begin{figure*}[p]
\centering
\includegraphics[width=\textwidth]{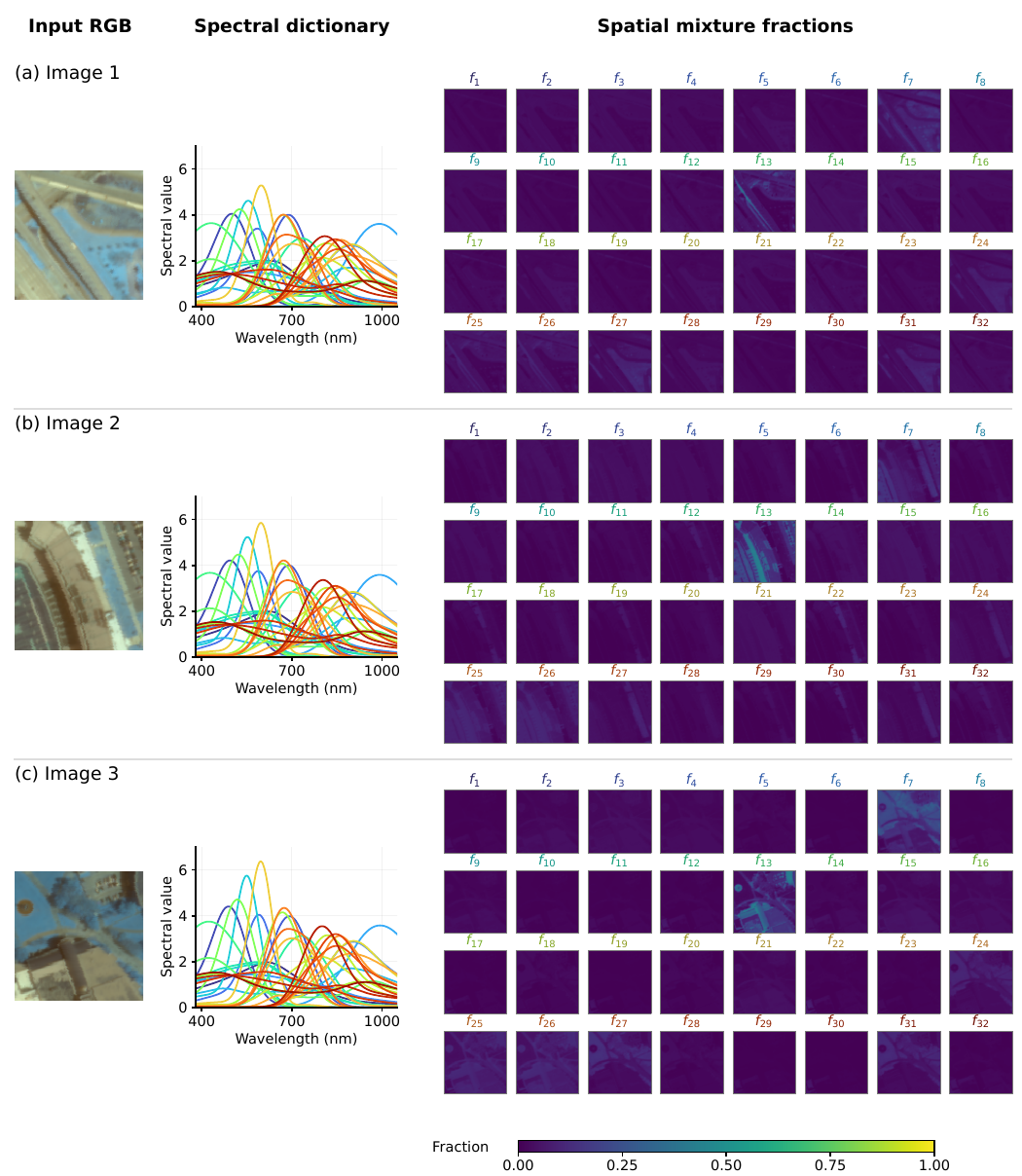}
\caption{Spectral dictionary and spatial mixture fractions with $M=32$ on the same three DFC2018 images. All 32 spectral primitives and their fraction maps are shown. Fraction maps use the common $[0,1]$ scale.}
\label{dictionary_size_m32}
\end{figure*}

Table~\ref{dictionary_size_results} provides the numerical results underlying the dictionary-size curves in the main paper (Fig.~\ref{dictionary_metrics}). Figs.~\ref{dictionary_size_examples}--\ref{dictionary_size_m32} show the spectral dictionaries and spatial mixture fractions for $M\in\{2,4,8,16,32\}$ on the same three DFC2018 images. For spectral primitive $s_m$, its spatial mixture fraction is defined as
\begin{equation}
 f_m(\p)=\frac{\sum_i\alpha_i(\p)\pi_{im}}{\sum_i\alpha_i(\p)}.
 \label{spatial_mixture_fraction}
\end{equation}
The fractions are defined where $\sum_i\alpha_i(\p)>0$ and set to zero otherwise. Unlike the primitive-level mixture weights $\pi_{im}$, $f_m(\p)$ measures the relative contribution of spectral primitive $m$ after spatial accumulation, with $\sum_m f_m(\p)=1$ wherever the total spatial weight is positive.

When $M=2$ (Fig.~\ref{dictionary_size_examples}), these spatial mixture fraction maps highlight complementary spatial regions. When $M=4$ (Fig.~\ref{dictionary_size_m4}), two maps maintain strong spatial contrast, while the remaining maps provide relatively uniform contributions. When $M=8$ (Fig.~\ref{dictionary_size_m8}), multiple maps exhibit similar spatial patterns, and the curves of certain spectral primitives overlap to a high degree. Therefore, increasing the number of primitives does not necessarily introduce significant spectral or spatial changes. When $M=16$ and $M=32$ (Figs.~\ref{dictionary_size_m16} and~\ref{dictionary_size_m32}), the dictionary contains more curves with varying peak positions and widths, yet several spatial mixture fraction maps still retain similar spatial patterns. Only a subset exhibits strong regional contrast.

\begin{table}[h]
	\centering
	\footnotesize
	\setlength{\tabcolsep}{2pt}
	\renewcommand{\arraystretch}{1.12}
	\caption{Effect of spectral dictionary size on DFC2018. \tablebest{Best} and \tablesecond{second-best} values are highlighted.}
	\label{dictionary_size_results}
	\begin{tabular*}{\columnwidth}{@{\extracolsep{\fill}}ccccc@{}}
		\toprule
		$M$ & RMSE$\downarrow$ & PSNR$\uparrow$ & SSIM$\uparrow$ & SAM$\downarrow$\\
		\midrule
		2 & \tablesecond{0.05856} & 27.9845 & 0.9101 & 12.5849\\
		4 & 0.05875 & 27.8755 & 0.9082 & \tablebest{12.3225}\\
		8 & 0.05891 & 27.8392 & 0.9090 & \tablesecond{12.4008}\\
		16 & 0.05874 & \tablesecond{28.0753} & \tablesecond{0.9107} & 12.6580\\
		32 & \tablebest{0.05840} & \tablebest{28.1135} & \tablebest{0.9170} & 12.5075\\
		\bottomrule
	\end{tabular*}
\end{table}

\end{document}